\pdfoutput=1

\documentclass[11pt]{article}
\usepackage{authblk}
\usepackage[]{acl}
\DeclareRobustCommand{\disambiguate}[3]{#2}

\usepackage{times}
\usepackage{latexsym}
\usepackage{enumitem}

\usepackage[T1]{fontenc}

\usepackage[utf8]{inputenc}

\usepackage{microtype}

\usepackage{graphicx}
\usepackage{hyperref}
\usepackage{csquotes}
\usepackage{xcolor}
\usepackage{adjustbox}
\usepackage{makecell}
\usepackage{multirow}
\usepackage{amsmath}
\usepackage{amssymb}
\usepackage{pifont} 
\title{Towards Enhancing Health Coaching Dialogue in Low-Resource Settings}

\author[1]{\bf Yue Zhou }
\author[1]{\bf Barbara Di Eugenio}
\author[1]{\bf Brian Ziebart}
\author[1]{\bf Lisa Sharp}
\author[1]{\authorcr \bf Bing Liu}
\author[2]{\bf Ben Gerber}
\author[1]{\bf Nikolaos Agadakos}
\author[1]{\bf Shweta Yadav}

\affil[1]{University of Illinois at Chicago, IL, USA}
\affil[2]{UMass Chan Medical School, MA, USA}
\affil[ ]{\texttt {\{yzhou232,bdieugen,bziebart,sharpl,liub,nagada2,shwetay\}@uic.edu}}
\affil[ ]{\texttt {\{ben.gerber\}@umassmed.edu}}

\begin{document}
\maketitle

\begin{abstract}

Health coaching helps patients identify and accomplish lifestyle-related goals, effectively improving the control of chronic diseases and mitigating mental health conditions. However, health coaching is cost-prohibitive due to its highly personalized and labor-intensive nature. In this paper, we propose to build a dialogue system that converses with the patients, helps them create and accomplish specific goals, and can address their emotions with empathy. However, building such a system is challenging since real-world health coaching datasets are limited and empathy is subtle. Thus, we propose a modularized health coaching dialogue system with simplified NLU and NLG frameworks combined with mechanism-conditioned empathetic response generation. Through automatic and human evaluation, we show that our system generates more empathetic, fluent, and coherent responses and outperforms the state-of-the-art in NLU tasks while requiring less annotation. We view our approach as a key step towards building automated and more accessible health coaching systems.  
\end{abstract}

\section{Introduction}

Health coaching is a patient-centered, motivational interviewing-based clinical practice that focuses on helping patients identify and accomplish personalized, lifestyle-related goals to improve health behaviors. It has been effective in improving the control of chronic conditions such as diabetes and cardiovascular disease and mitigating mental health conditions such as anxiety and depression~\citep{hc4butterworth2006effect,hc1ghorob2013supplement,hc3kivela2014effects,hc2thom2016qualitative}. Health coaching can be particularly beneficial to low-socioeconomic status (SES) populations who disproportionately suffer physical and mental disease burdens~\citep{ses2thackeray2004disparities,ses1kangovi2014challenges}. Yet, it is invariably cost-prohibitive for these populations due to its highly personalized and labor-intensive nature.

\begin{figure}[!t]
    \centering 
    \includegraphics[width=\columnwidth]{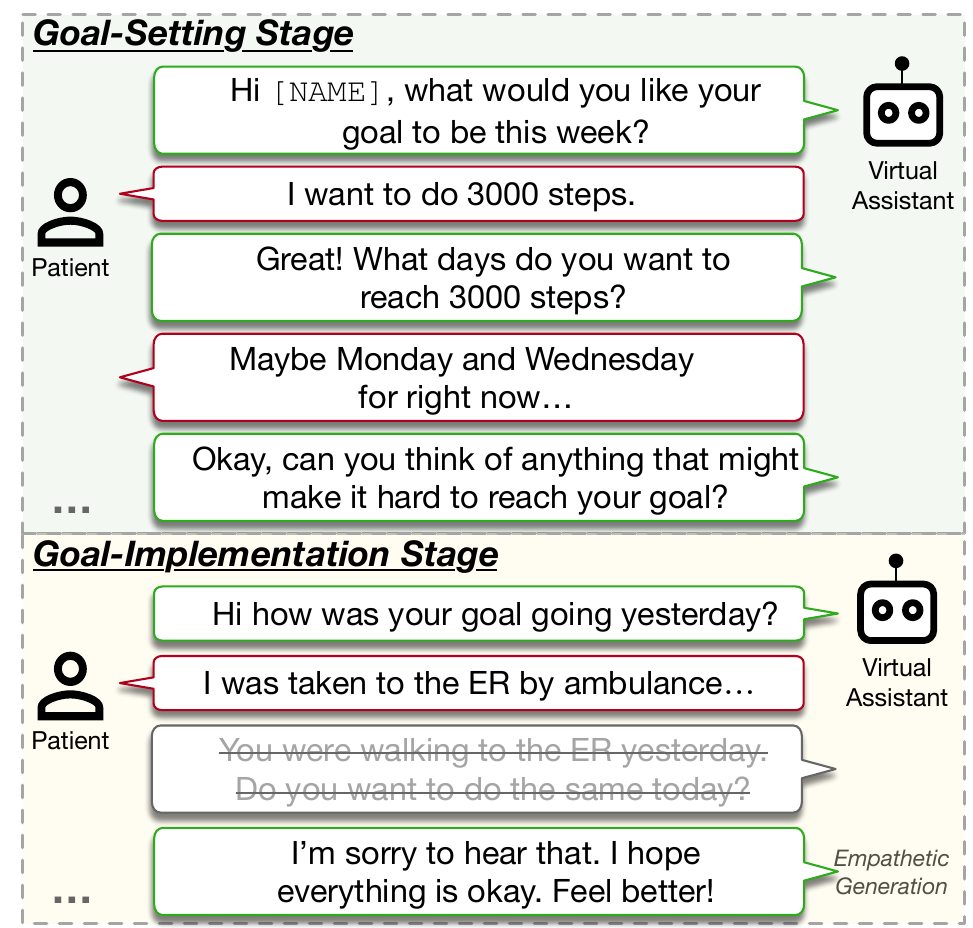}
    \caption{A health coaching dialogue scenario in our dialogue system. It starts with a goal-setting stage where the coach discusses a realistic goal with the patient. After the goal is settled, the coach follows up on the patient's progress and maintains patient engagement. Understanding the patient's emotion cues and responding empathetically is also crucial in such a scenario; the struck-out response, generated by a na\"ive sequence-to-sequence model, is inappropriate without the capability of understanding and modeling empathy.}
    \label{fig:teaser}
\end{figure}

Facilitating conversations in healthcare settings between the participants via language techniques and text messages~\citep{sms1aguilera2011text,sms2fitzpatrick2017delivering} has the potential to improve the efficacy of health coaching while reducing the cost. However, the interactions between the dialogue systems and patients are either scripted or with limited natural language understanding (NLU) or generation (NLG) capabilities~\citep{sms5kocielnik2018reflection,sms3chaix2020vik,sms4mohan2020designing}. In our previous work \citet{gupta1-etal-2020-human,gupta22020goal,gupta32021summarizing}, we collected real-world health coaching conversation datasets and focused on the NLU components of the dialogue. However,  no existing health coaching dialogue system supports natural language conversations between the patient and a coach agent.

In this paper, we propose to build a dialogue system that converses with patients and helps them create and accomplish specific goals for physical activities based on the health coaching dataset we previously collected~\citep{gupta22020goal}. We want the system to emulate the health coaching process, which starts with a goal-setting stage where the coach discusses creating a S.M.A.R.T goal with the patient, namely a goal that is specific, measurable, achievable, relevant, and time-bound \cite{doran1981there}.
Once the goal is settled, the coach agent would follow up on the patient's progress and maintain patient engagement. In addition, we focus on developing the system to understand and address the patient's emotions and respond empathetically, which is crucial for better procedure outcomes in healthcare settings~\citep{emp-import1levinson2000study,emp-import2moudatsou2020role}. A health coaching dialogue example is shown in Figure~\ref{fig:teaser}. 

Building such a dialogue system is challenging: First, the real-world health coaching dataset is limited in size and annotation, which restricts not only end-to-end but also modularized approaches, especially with no dialogue states annotated. Additional annotation is also resource-intensive. Second, generating empathetic responses is subtle, which may require incorporating external empathetic knowledge since cases of empathetic responses in our health coaching dataset are rare. 

To address these challenges, we propose a modularized task-oriented health coaching dialogue system with a simplified architecture that requires fewer annotations. The system contains an NLU module, an \texorpdfstring{NLG\textsubscript{hc}}{NLG hc} module, and an \texorpdfstring{NLG\textsubscript{emp}}{NLG emp} module. The NLU module aims to keep track of the goal attributes as simplified belief states. 
The \texorpdfstring{NLG\textsubscript{hc}}{NLG hc} module takes as input the current dialogue context, the belief states, and the coaching stage to generate coaching utterances. Finally, for the \texorpdfstring{NLG\textsubscript{emp}}{NLG emp} module, we build an emotion cue detector and mechanism-conditioned empathy generator to facilitate empathetic response generation.


To evaluate our approach, we combine automatic evaluation with expert-based human evaluation. Our experimental results show that our NLU module outperforms the state-of-the-art~\citep{gupta1-etal-2020-human,gupta22020goal,gupta32021summarizing} by > 10\% in F1-score in the slot-filling and > 7\% in semantic frame correctness in the offline goal attributes tracking task, which also enables updating goal information online at every dialogue turn. Moreover, the experiments demonstrate that our dialogue generation achieved best performance compared to baseline methods in terms of coherence, fluency, and empathy. Finally, through a pilot human evaluation, our model's generation is preferred by the health coaches as concerns  coherence and empathy. 

The contributions of this work are: (1) We propose to build an efficient modularized health coaching dialogue system that helps patients create and accomplish specific goals, with a simplified NLU and NLG framework combined with mechanism-conditioned empathetic response generation. (2) Our system outperforms the state-of-the-art in NLU tasks while significantly reducing the annotation workload. (3) Through automatic and human evaluation, we show our system generates more coherent and empathetic responses, which can provide suggestions for health coaches and improve coaching efficiency.
\section{Related Work}
\begin{itemize}[]
\item \textbf{Conversational Agents in Healthcare.} Conversational agents have been explored to improve the efficacy and scalability of the interactions between healthcare professionals and patients. For instance, chatbots in different healthcare settings, including chronic disease monitoring~\citep{sms3chaix2020vik}, cognitive behavior therapy~\citep{sms2fitzpatrick2017delivering}, and physical activity promotion~\citep{sms4mohan2020designing,sms5kocielnik2018reflection}. However, they are limited in natural language understanding and generation capabilities. More sophisticated approaches have been proposed in mental health counseling~\citep{consolalthoff-etal-2016-large,consol2shen-etal-2020-counseling}. In our previous work \citet{gupta1-etal-2020-human,gupta22020goal,gupta32021summarizing}, we collected real-world health coaching conversation datasets and focused on the NLU components of the dialogue which summarize weekly goals to support health coaches.  

\item \textbf{Task-Oriented Dialogue.}
Traditional task-oriented dialogue systems are modularized \citep{jokinen09}. They consist of an NLU component for understanding the user intent and recording the dialogue states and an NLG component for policy management and response generation~\citep{tod1williams2016dialog,tod2budzianowski-etal-2018-multiwoz,tod3mrkvsic2016neural,tod4wen-etal-2015-semantically}. However, approaches have shifted towards end-to-end architectures to reduce human effort and error propagation between modules~\citep{tode2ebordes2016learning,tode2e2wen2016network}. Recently, training an end-to-end system as a sequence prediction problem leveraging the causal language models has delivered promising results~\citep{tode2e3hosseini2020simple,tode2e4peng2020soloist}. 

\item \textbf{Empathetic Data for Conversations.}
Empathetic interaction is a key to better task outcomes in conversations. Recently, empathetic data and approaches have been proposed to facilitate empathetic conversations in open-domain and healthcare settings. Babytalk \cite{hunter-ecai2008,mahamood-inlg2011} is an earlier system that summarizes  neonatal intensive care patient data for different types of users, and provides affective cues for the parents.  \citet{emp-ed-rashkin-etal-2019-towards} proposed an open-domain empathetic dialogue dataset (ED), with each dialogue grounded in an emotional context. \citet{empmoviewelivita-etal-2021-large} extended from ED and proposed a large-scale silver-standard empathetic dialogue data based on movie scripts. \citet{emp-type-data-sharma-etal-2020-computational,empRL10.1145/3442381.3450097} proposed an empathetic dataset in mental health support settings, with the communication mechanisms and corresponding strength of empathy annotated. 
\end{itemize}

\section{Health Coaching Dataset} 

\label{hc-dataset}

We first briefly describe our dataset, that motivates the approach that we are proposing; full details can be found in
\citet{gupta1-etal-2020-human}. We collected two datasets of health coaching dialogues between patients and coaches via text messages. Dataset~1 and~2 contain 28 and 30 patients, with each patient coached for four and eight weeks, respectively. Each week the health coach converses with the patient to create a  physical activity S.M.A.R.T. goal and then follows up on the patient's progress. The two datasets contain 336 weeks of dialogues with 21.4 average turns per week. 

We defined ten slots for the goal's attributes (types of activity, amount, time, days, location, duration, and the confidence score for the activity) (the Appendix contains examples of all slots). We used a stage-phase schema for additional turn-level annotation describing how the health coaching dialogue unfolds. We defined two stages: the goal-setting stage and the goal-implementation stage. Each stage includes a set of phases, such as goal identification, negotiation, and follow-up. Each turn can belong to a certain stage-phase combination. Later, we added dialogue act annotations~\citep{gupta32021summarizing}, consisting of  12 domain-independent dialogue acts,  following the ISO-standard by~\citet{isoactmezza-etal-2018-iso}.

Since dataset~1 was collected before dataset~2, the manual annotation was developed on dataset~1, and used to develop our NLU component, that was then tested on dataset~2. Hence, dataset~1 is fully annotated  for  slot-value spans, goals, stages and phases.
In dataset~2, only three patients are annotated for slot-value spans and phases, and 15 weeks for goals; the whole dataset is annotated for stages. As far as dialogue acts are concerned, they are only available for 15 weeks of dialogues in dataset 1\footnote{The datasets with annotations are available at \url{https://github.com/uic-nlp-lab/virtualcoachdata}.}.

\section{Methods} 

In this section, we begin by first providing a brief workflow of our proposed health coaching dialogue system and then describing the model architecture in details.


\subsection{Health Coaching Dialogue System}


A health coaching dialogue can be framed into stages: \textbf{(1)} Starting with the goal-setting stage, the coach helps the patient create a specific goal whose attributes can be represented by a list of slot values\footnote{We provide a complete list of slots and corresponding examples in the Appendix.} (\textit{e.g., Activity = Walk; Amount = 3000 steps; Days = Mon-Fri}), which retains a task-oriented nature. However, it is much more complex than the prevailing task-oriented services, since \textbf{(2)} after the goal is initialized, the coach needs to follow up on the patient's progress and maintain patient engagement (e.g., checking in, sending reminders, revising the goal, and providing encouragement on patient): this is  the goal implementation stage. A health coaching dialogue contains multiple turns. We denote the stage that the turn $t$ belongs to as $S_t$.

Leveraging the task-oriented dialogue framework, we use belief state $B_t$, a list of slot-value pairs, to record the goal attributes in a turn $t$. 
An \textbf{NLU} module is used to infer $B_t$ by considering the earlier patient utterances $U_{<t}$, current patient utterances $U_t$, and earlier system responses $R_{<t}$ as input to the module. Formally:
\begin{equation}
B_t = \textrm{NLU}([U_0, R_1, U_1,R_2,...,U_t])
\end{equation}
 $B_t$ summarizes the goal from the dialogue history and will be used for conditional response generation and lexicalizing the generated response.
 
 Then, given the dialogue context $C_t$ in turn $t$, consisting of the previous two turns [$R_{t-1}$,$U_{t-1}$], the stage of the previous turn $S_{t-1}$, and $B_{t}$, we build a sequence classifier to predict the stage $S_t$:
\begin{equation}
S_t = \textrm{Seq2Label}_{hc}([C_t,B_t,S_{t-1}])
\label{the 2}
\end{equation}
Finally, a delexicalized response $R_t$ is generated    
via a Seq2Seq neural network given $C_t$, $B_t$, and $S_t$ concatenated as a single sequence:
\begin{equation}
R_t = \textrm{Seq2Seq}_{hc}([C_t,B_t,S_t])
\label{the 3}
\end{equation}
The response can be lexicalized into human readable text using the belief state $B_t$.
Equations \ref{the 2} and \ref{the 3} constitute the \textbf{\texorpdfstring{NLG\textsubscript{hc}}{NLG hc}} module, where we explore replacing the fine-grained action prediction with the coarse stage prediction as proximal dialogue management. 

Although \textbf{\texorpdfstring{NLG\textsubscript{hc}}{NLG hc}} could learn limited empathetic response patterns such as \textit{"sorry to hear that."} from the health coaching data, it does not retain prior empathetic knowledge and explicitly model empathy. However, showing a caring attitude and being empathetic to patients' emotional cues are crucial to patient activation and engagement, leading to better task outcomes. To facilitate the empathetic capability of the system, we build the \textbf{\texorpdfstring{NLG\textsubscript{emp}}{NLG emp}} module, where the empathetic response $\tilde{R}_t$ is generated conditioned on the patient's previous turn utterance $U_{t-1}$ and communication mechanism signals $M$:

\begin{equation}
\tilde{R}_t = \textrm{Seq2Seq}_{emp}([U_{t-1},M])
\label{emp 4}
\end{equation}
We follow~\citet{emp-type-data-sharma-etal-2020-computational} and consider three communication mechanisms for empathy: \textit{`Emotional Reactions'}, \textit{`Interpretations'}, and \textit{`Explorations'}. Emotional reaction expresses direct emotions (e.g., compassion) to show empathy, such as \textit{"I would be very worried."} Interpretation communicates an understanding of the speaker's experience, such as \textit{"I know anxiety is scary."} Exploration expresses empathy by exploring the speaker's feelings, such as \textit{"What happened? How come?"} An empathetic response can be realized through multiple communication mechanisms. Thus, $M$ contains one or more of the three special tokens \texttt{[EMOR]}, \texttt{[INTERP]}, and \texttt{[EXPLOR]}, representing the three communication mechanisms. We seek to use the mechanism signals to control the style of the generated empathetic response, making it flexible and appropriate for health coaching scenarios.

In addition, an emotion cue detector is built to support empathetic generation. We predict the current emotion signal $E_{t}$ given the patient's previous utterance $U_{t-1}$. Then $\tilde{R}_t$ is generated when a strong emotion cue is detected with the probability of certain types of emotion greater than a predefined threshold, \textit{i.e.},  $p(E_{t}|U_{t-1})>\tau$, obtained from development set performance.

\begin{figure*}[!htb]
    \centering 
    \includegraphics[width=0.8\textwidth]{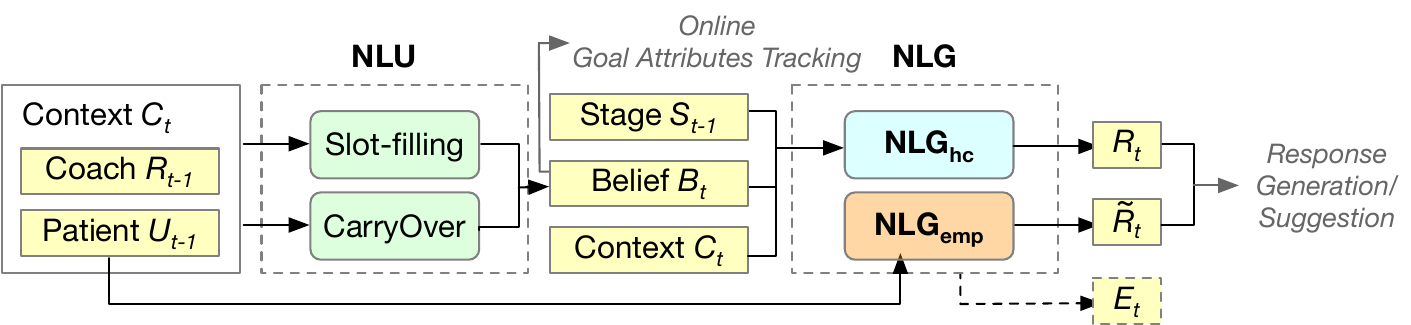}
    \caption{The framework of our health coaching dialogue system. The NLU module consisting of the slot-filling and carryover model reads the dialogue and infers belief state $B_t$. The \texorpdfstring{NLG\textsubscript{hc}}{NLG hc} module takes as input the stage $S_{t-1}$, belief state $B_t$, and context $C_t$ to generate response $R_t$. The \texorpdfstring{NLG\textsubscript{emp}}{NLG emp} handles the cases where empathy are required and outputs empathetic response $\tilde{R}_t$ and emotion signal $E_t$.}
    \label{fig:of1}
\end{figure*}

The overall framework of our model is illustrated in Figure~\ref{fig:of1}.

\subsection{Model Architectures}

In this section, we describe the detailed model architecture for each of the NLU, \texorpdfstring{NLG\textsubscript{hc}}{NLG hc}, and \texorpdfstring{NLG\textsubscript{emp}}{NLG emp} module. 

\subsubsection{The NLU Module}

The NLU module records the current state of the slot values. It consists of a neural slot-filling model that extracts the slot fillers from each utterance and a carryover classifier to determine if the value of a slot should be copied from the previous state. 

\paragraph{Neural Slot-Filling} A neural slot-filling model maps the sentence representation to a sequence of BIO labels (the beginning
\textit{(B)} and inside \textit{(I)} of each slot label, and outside \textit{(O)} for others), often combined with sentence-level classifications (e.g., domain, intent). Following~\citet{jointbertchen2019bert}, we use BERT~\citep{devlin-etal-2019-bert} as the model backbone, the \texttt{[CLS]} token of the sentence for sentence-level classification, and the final hidden state of the first sub-token of each word at position $n$ for BIO labeling. The network is trained by maximizing the conditional probability:
\begin{equation}
p\left(y^{c},\boldsymbol{y^{s}} \mid \boldsymbol{x}\right) = p\left(y^{c} \mid \boldsymbol{x}\right) \prod_{n = 1}^{N} p\left(y^s_n \mid \boldsymbol{x}\right)
\label{sf1}
\end{equation}
Where $y^c$, $y^c_n$ are softmax probabilities for sentence labels and BIO labels of word $n$, with  $\boldsymbol{x}$ being the sequence of word tokens.

\paragraph{Carryover Classifier} The carryover classifier determines if the value of a slot should be copied from the previous state or updated with the new instance seen in the current utterance, which outputs a binary value for each slot. 

Following the work of ~\citet{carryovergao-etal-2019-dialog}, we developed our slot carryover classifier based on the current dialogue context. Our proposed carryover classifier, however, differs from~\citet{carryovergao-etal-2019-dialog}, in terms of: (1) we design the classifier as a separate model rather than a component in an end-to-end belief state tracking architecture; (2) it benefits from the results of the slot-filling model and only makes a prediction when there is a collision between the existing value and the new for a given slot. Such a design takes into account that: (1) dialogue datasets in healthcare are limited in size, making end-to-end training difficult; (2) these datasets can also be limited in annotations. Our model can facilitate annotation since the annotators can work with the slot-filling model to only examine the lines where the value collision occurs, enabling more efficient labeling for belief states in a carryover fashion.

We use BERT to encode the contexts and utilize the hidden state associated with the \texttt{[CLS]} token to represent the current context $C_t$ and maximize $p(y_t|C_t)$ via fine-tuning BERT with cross entropy minimization:
\begin{align*} 
h_t = \textrm{BERT}(C_t)\textrm{[CLS]} \\
y_t = \textrm{Softmax}(Wh_t+b)
\end{align*}
where $y_t \in [0,1]^{N_s}$, $N_s$ is the number of slots.

During inference, the slot-filling model extracts the slot fillers at each turn. The carryover classifier determines if the value should be copied from the previous state or updated with the new instance given dialogue context when there is a value conflict. NLU enables an update for the belief state $B_t$ at each turn, \textit{i.e.,} providing online goal attributes tracking. 
\subsubsection{The \texorpdfstring{NLG\textsubscript{hc}}{NLG hc} Module} The dominant modularized approach for task-oriented dialogue often consists of belief state tracking, dialogue policy, and language generation. It requires dedicated modeling and annotations for dialogue acts. 
However, annotating acts is resource-intensive in healthcare settings. 

An alternative is using stages, where each stage restricts a set of possible actions in health coaching dialogue and other healthcare conversations, such as clinical motivational interviews and patient encounters. For example, in health coaching, sending reminders and encouragement on progress can only appear in the goal implementation stage when the goal has already been created. Another example is the SOAP (Subjective, Objective, Assessment, and Plan) structure~\cite{soap-podder2021soap} of patient encounters, in which discussing a patient's chief complaint occurs in the Subjective part, while diagnosis is discussed in the Assessment part.

\begin{figure*}[!htb]
    \centering 
    \includegraphics[width=0.8\textwidth]{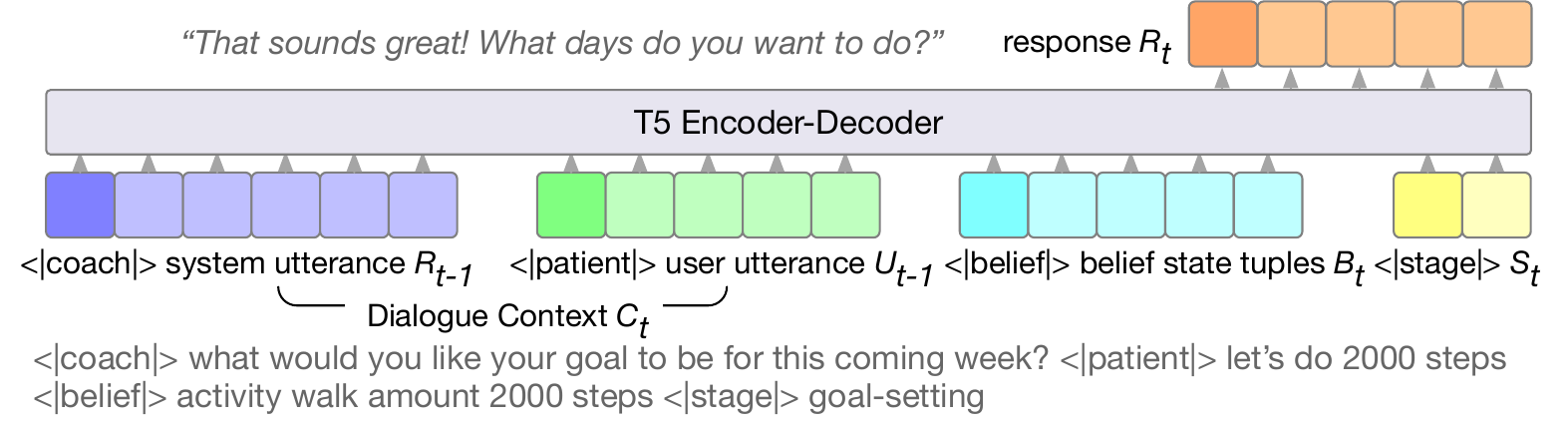}
    \caption{Model architecture of generating $R_t$ in \texorpdfstring{NLG\textsubscript{hc}}{NLG hc}. The context $C_t$, belief tokens $B_t$, and the predicted stage $S_t$ are concatenated as a single input sequence, training with a T5 encoder-decoder model.}
    \label{fig:nlg01}
\end{figure*}

Thus, we explore the possibility of using stages that contain coarse and fuzzy act information to guide dialogue generation instead of fine-grained act annotations. Concretely, we use T5~\citep{t5-raffel2019exploring} to jointly model $\textrm{Seq2Label}_{hc}$~(\textit{cf.}\ref{the 2}) and $\textrm{Seq2Seq}_{hc}$~(\textit{cf.}\ref{the 3}) for stage $S_t$ prediction and response $R_t$ generation as a multi-task approach. The input is a single sequence by concatenating the local dialogue context, current belief state, and stage tokens. For example, the input for $\textrm{Seq2Seq}_{hc}$ is the concatenation of $C_t$, $B_t$, and $S_{t}$, separated by delimiter tokens, which is mapped to the target response sequence, as shown in Figure~\ref{fig:nlg01}. The model is encoder-decoder based, trained with a maximum likelihood objective with a different prefix prepending to the input corresponding to each task.

\subsubsection{The \texorpdfstring{NLG\textsubscript{emp}}{NLG emp} Module}

\paragraph{Empathetic Response Generation}

We fine-tune GPT2~\cite{gpt2-radford2019language} for empathetic response generation. At training time, we concatenate the user utterance, the corresponding empathetic response, and the communication mechanisms used in the response, separated by delimiter tokens, as one single training sequence $\boldsymbol{x}$. We want GPT2 to learn to model the joint probability $p(\boldsymbol{x})$ by fine-tuning on empathetic data. 
During inference time, the model takes as input the user utterance and the communication mechanisms and generates one token at a time for empathetic response. 

The following shows an example of such a training sequence: 

\begin{displayquote}
 \texttt{<|bos|>} \textcolor{black}{\texttt{[EMOR]}} \textcolor{blue}{I was so exhausted yesterday.} \textcolor{black}{\texttt{<|sep|>}} \textcolor{orange}{That's understandable. Take some rest!} \texttt{<|eos|>}
\end{displayquote}
where \texttt{<|bos|>} and \texttt{<|eos|>} are the delimiters for beginning of sequence and end of sequence, \texttt{[EMOR]} stands for the communication mechanism \textit{Emotional Reaction}, and \texttt{<|sep|>} separates the \textcolor{blue}{user utterance} and \textcolor{orange}{empathetic response}
. During inference, the model starts to generate tokens after \texttt{<|sep|>}.
\paragraph{Emotional Cue Detection}

To support empathetic response, we build a BERT-based multi-class classifier to predict the emotion signals of the patient's utterance. In this work, we use the softmax probability for the predicted label to determine if the empathetic response needs to be generated. However, the predicted emotion labels also facilitate future sentiment analysis work in health coaching dialogue.

\section{Experiments}

This section includes datasets description, evaluation metrics, experimental results, and qualitative analysis.

\subsection{Dataset Details}

We conducted our experiments on the following benchmark datasets:

\paragraph{Health Coaching Datasets}

These are the datasets we described earlier in Section~\ref{hc-dataset}. Following~\citet{gupta22020goal}, we use dataset~1 for training/development and dataset~2 for testing  for both NLU and \texorpdfstring{NLG\textsubscript{hc}}{NLG hc} to ensure comparability. 

\paragraph{Data Augmentation for Slot-filling}
We only consider the utterances that contain at least one slot-value span for slot-filling. This results in 955:105:205 data points for training, development, and testing. To alleviate data scarcity, we use a na\"ive data augmentation method to synthesize training examples that (1) first randomly replaces the value with one possible alternative for each slot; (2) then rephrases the modified sentence with a pre-trained paraphrase model\footnote{A fine-tuned PEGASUS~\citep{parazhang2019pegasus} on PAWS~\citep{paws2019naacl}}. 

\paragraph{Empathetic Datasets}

We utilized two benchmark empathetic dialogue datasets:  \textit{(i)} \textsc{Empathetic}\textsc{Dialogues} (ED)~\citep{emp-ed-rashkin-etal-2019-towards} and \textit{(ii)} EPITOME. \textsc{Empathetic}\textsc{Dialogues} (ED) is a open-domain empathetic dialogue dataset containing 24,850 dialogues, with each dialogue grounded in one of 32 emotional context (e.g., proud, apprehensive, confident, guilty). EPITOME \citep{emp-type-data-sharma-etal-2020-computational} is an empathetic dataset in mental health support settings. The dataset consists of 3k posts annotated with respect to the communication mechanisms (\textit{Emotional Reaction, Exploration, Interpretation}) and level of empathy (from 0 to 2, representing no empathy, weak and strong empathy, respectively) in each mechanism.

To facilitate empathetic generation conditioned on the communication mechanisms, we first build a weak multi-label classifier for predicting the types of communication mechanisms based on the 3k posts in \citet{emp-type-data-sharma-etal-2020-computational} (achieving a hamming loss~\citep{hamminglosstsoumakas2007multi} of $\sim$0.16). Then we apply the classifier on the ED dataset to get communication mechanism labels for each instance. We use their default train/dev/test split. Finally, $N = 64$ empathetic cases in the health coaching dataset are used for few-shot fine-tuning the trained empathetic generation model. 


\subsection{Evaluation Metrics}

Depending upon the task, we use the following evaluation metrics:

\begin{enumerate}[noitemsep]
   \item Slot-filling: Precision, Recall, and F1-score.
    
   \item Dialogue State (Goal Attribute) Tracking: partial/complete match and goal correctness@$k$ following~\citet{gupta22020goal}:
    \begin{description}
     \item[Correctness@$k$:] Computes the percentage of correctness over all the predicted goals of each week. If at least $k$ attributes are predicted correctly, the goal is regarded as correct. 
     
     \end{description}
    \item Dialogue Generation: BLEU~\citep{bleu-papineni-etal-2002-bleu}, BertScore~\citep{bert-score}, Perplexity (PPL), and Empathy Score.
     \begin{description}
     \item[Perplexity:] We measure fluency as perplexity (PPL) of the generated response using a pre-trained GPT2 model that has not been fine-tuned for this task, following previous work~\citep{ppl-ma-etal-2020-powertransformer,emp-rewriting10.1145/3442381.3450097}.
     \item[Empathy Score:] We train a standard text regression model based on BERT using the response posts and corresponding level of empathy scores in  \citet{emp-type-data-sharma-etal-2020-computational} (achieving an RMSE of $\sim$0.57). We use this model to measure the empathy in the generated outputs.  
     \end{description}
\end{enumerate}




For detailed descriptions of metrics, training, including model parameters, selection, and supplementary analysis, please see the Appendix.

\begin{table}[h]
\centering
\begin{adjustbox}{width=0.9\columnwidth}

\begin{tabular}{lllll}
\hline
        System                   & Slot R         & Slot P         & Slot F1        & PAcc     \\ \hline
\citet{gupta22020goal}          & 0.806          & 0.808          & 0.790          & 0.801          \\
+Phase                 & 0.899          & 0.837          & 0.867          & 0.779          \\
+StartPhase            & 0.910          & 0.847          & 0.877          & \textbf{0.835} \\
+StartPhase+Aug        & \textbf{0.926} & \textbf{0.879} & \textbf{0.902} & 0.817          \\
Slot Only+Aug & 0.904          & 0.876          & 0.890          & -   \\     
\hline
\end{tabular}

\end{adjustbox}
\caption{Evaluation on slot-filling with ablations. \textit{Aug}: using data augmentation; \textit{StartPhase}: Jointly predicting the phase of the sentence only if it is the beginning sentence of the phase.
}
\label{tab:sf}
\end{table}

\begin{table*}[h]
\centering
\begin{adjustbox}{width=\textwidth}

\begin{tabular}{llllllllllllllll}
\hline
           & \multicolumn{5}{c}{Dataset 1 Backward}                                          & \multicolumn{5}{c}{Dataset 1 Forward}                                           & \multicolumn{5}{c}{Dataset 2}                                                 \\ \hline
           & Complete       & Partial        & @k=10         & @k=9          & @k=8          & Complete       & Partial        & @k=10         & @k=9          & @k=8          & Complete     & Partial        & @k=10         & @k=9          & @k=8          \\ \hline
\citet{gupta22020goal}   & 0.77           & 0.81           & 13.1          & 39.3          & \textbf{65.4} & 0.81           & 0.85           & 15.0          & 44.9          & 68.2          & -            & -              & -             & -             & -             \\
ourSF+Rule & \textbf{0.808} & \textbf{0.858} & \textbf{31.6} & \textbf{42.1} & 63.2          & 0.831          & 0.863          & 21.7          & 69.6          & 73.9          & 0.767        & 0.787          & 13.6          & 18.2          & 40.9          \\
ourNLU     & 0.792          & 0.825          & 15.8          & 36.8          & 63.2          & \textbf{0.838} & \textbf{0.869} & \textbf{21.7} & \textbf{69.6} & \textbf{78.3} & \textbf{0.8} & \textbf{0.813} & \textbf{13.6} & \textbf{27.3} & \textbf{63.6} \\
\hline
\end{tabular}
\end{adjustbox}
\caption{Evaluation on offline goal attributes tracking with complete/partial match and goal correctness@$k$.
}
\label{tab:goal-summ}
\end{table*}

\begin{table*}[h]
\begin{adjustbox}{width=\textwidth}
\begin{tabular}{lll}
\textbf{Input}                                                        & \textbf{Model}   & 
\textbf{Generated Response}                                                     \\
\hline\hline
Coach: What would you like your goal to be this week?         & Ours    & Sounds good, \textcolor{teal}{which days were in your mind?}                             \\
Patient: I want to \textcolor{blue}{walk 30 min a day between 6am to 8am}.       & IR      & And \textcolor{teal}{what days} would you like to walk \textcolor{red}{the track around neighborhood}?    \\
                                                             & Seq2Seq & \textcolor{teal}{How many days} do you want to \textcolor{red}{reach} this week?                          \\
                                                             \hline
Coach: Good morning! How is your goal for this week going?    & Ours    & \texttt{[EMOR]} Oh no, \textcolor{teal}{I hope you are okay.}                                       \\
Patient: I'm sorry I didn't go to work today I have \textcolor{blue}{a massive} &         & \texttt{[EXPLOR]} Oh geez, sorry to hear that. \textcolor{teal}{Are you feeling better? }           \\
\textcolor{blue}{migraine headache}.                                          &       & \texttt{[INTERP]} \textcolor{teal}{I've had this experience before. Sometimes it really hits you.}                                             \\
                                           & IR      & \textcolor{red}{Check your steps.}                                              \\
                                                             & Seq2Seq & Have you \textcolor{red}{been busy with the migraine} this week and how did it turn out? \\
                                                             \hline\hline
                                                             Coach: \textcolor{blue}{Don't forget to} take a deep breath - breathing is important.         & Ours     & \textcolor{red}{Don't forget to} take a deep breath - breathing is important.                            \\
Patient: You made me smile. Thanks.       &   \ding{55} Repeat   &    \\
\hline
Coach: Hi when you get a chance can you sync your fitbit?         & Ours     & \textcolor{red}{That's great news.} I hope you \textcolor{teal}{don't get into a long trip there}.                            \\
\hspace{3em} I only see your steps since last Friday. Thanks. &    \ding{55} Focus  &                            \\
Patient: Sorry \textcolor{blue}{been in Texas}, just \textcolor{blue}{made it back to Chicago}.    &      &    \\
                                                             \hline
\end{tabular}
\end{adjustbox}
\caption{Qualitative examples of generated responses in empathetic and non-empathetic scenarios, combined with error analysis.}
\label{tab:qa}
\end{table*}

\subsection{Results}

\subsubsection{Goal Attributes Tracking}

Table~\ref{tab:sf} shows the performance of slot-filling and phase prediction compared to the previous model. Jointly predicting the phase of the sentence only if it is at the beginning of the phase, combined with data augmentation (+StartPhase+Aug), achieved the best performance for slot-filling, outperforming the state-of-the-art by 11.2\% in F1. However, modeling without phases suffices for slot-filling while reducing the annotation cost. As such, we adopt the no-joint model for downstream tasks. Our experimental result also shows the carryover classifier achieved a F1-score of 0.88 using only the dialogue context. We investigate whether dialogue act and phase labels can improve carryover classifier, however they barely contribute to the model performance\footnote{See Appendix for detailed results.}.

In previous work, we extracted goals at two critical points for each week to evaluate offline goal tracking: one at the end of the goal-setting stage (forward) and the other at the end of the goal-implementation stage (backward). The forward and backward goals can be different since the patient may encounter barriers, and the goal can be revised in the implementation stage. We also proposed a rule-based approach to update the slot values, which simply records the last mention of the value for each slot except for certain conditions. In this paper, we compared our model with previous work and a combination of our slot-filling model with the previous rule-based approach (our SF+Rule). Table~\ref{tab:goal-summ} shows the performance for goal attributes tracking of our NLU module compared with previous work. For dataset 1 backward goals, our SF+Rule achieved the best performance resulting from more accurate slot-filling. We observe that in dataset 1, the coach tends to summarize the goal to the patients at the end of each week, which benefits the rule-based approach. Nonetheless, our NLU module outperforms previous work in all evaluation metrics on dataset 1 forward and dataset~2; it  also enables goal tracking at each turn (online).

\begin{table}[h]
\centering
\begin{adjustbox}{width=0.85\columnwidth}

\begin{tabular}{lllll}
\hline
Model   & BLEU  ($\uparrow$)        & PPL ($\downarrow$)           & BertS F1 ($\uparrow$)  & EmpS ($\uparrow$)      \\ \hline
IR      & 0.194          & 24.2          & 0.853          & -0.179          \\
Seq2Seq & 0.242          & 16.26         & 0.863          & +0.073          \\
\hspace{0.2cm}+Acts & 0.235          & 16.2         & 0.861          & -0.065          \\

OURS    & \textbf{0.251} & \textbf{15.6} & \textbf{0.872} & \textbf{+0.256}\\
\hline
\end{tabular}

\end{adjustbox}
\caption{Evaluation on dialogue generation with automatic metrics. BLEU: Average of BLEU-1,-2,-3,-4. EmpS: Computed by the difference of the empathetic score between the output and the ground truth ($\sim$ 0.978).}
\label{tab:gene}

\end{table}

\subsubsection{Dialogue Generation}

Since our previous work did not include  a generation model, we compare our dialogue generation with three baselines: (1) Retrieval-based model. We use BERT to encode the current belief state and dialogue context as query $h_q$, and encode the response as $h_r$. We fine tune the BERT models to retrieve the response that maximizes $h_q\cdot{h_r}$; (2) A seq2seq neural network that maps the belief state and dialogue context to the output without stages. (3) using dialogue acts instead of stages.

Table~\ref{tab:gene} shows that our response generation outperforms all baselines in all metrics. The predicted stage information (achieved an accuracy of $\sim$0.92) provides meaningful signals for dialogue generation. In addition, by incorporating external empathetic knowledge, our model achieved +0.256 average improvement in empathy. 
Including dialogue acts does not improve the performance compared to the seq2seq model. 
This is not unexpected because labels are imbalanced and only available for 15 weeks of data. Computationally labeling dialogue acts also leads to large error propagation. 
The emotional cue detection (32-category classification) achieved an accuracy of $\sim$0.58. Based on development set observation, we set $\tau$ to be 0.7 and generate empathetic response when the cumulative probability of the top-2 predicted emotional labels is greater than $\tau$.

\paragraph{Human Evaluation.} We performed an expert-based human evaluation on coherence and empathy through A/B testing. We ask two health coaches to compare outputs from our model against other baselines given the same input and choose (a) the response which is more empathetic; (b) the response which is more coherent. Among the collected 43 examples, our model's outputs have a $\sim$$71\%$ preference for empathy and $\sim$$55\%$ preference for coherence over other baselines.

\subsection{Qualitative Analysis}
We present examples of our health coaching dialogue generation and mechanism-conditioned empathetic generation in Table~\ref{tab:qa}. In the first case, where empathy is not required, all three models choose to inquiry for more information of the goal. However, the fixed, retrieval-based response can contain context-irrelevant tokens (\textit{"..the track around neighborhood"} ) and the Seq2Seq response is relatively unnatural.  
In the second case, given the cues including \textit{"..a massive migraine headache.."}, our model can generate corresponding empathetic responses given different communication mechanisms (e.g., \texttt{[EMOR]} $\rightarrow$ \textit{"Oh no, I hope you are okay."}). In contrast, the retrieval-based and Seq2Seq model failed to response empathetically. Particularly, the Seq2Seq model failed to distinguish a symptom (\textit{i.e., "migraine headache"}) from a physical activity or goal. Finally, we present two incoherent generation examples by our model. In the first example, our system misinterprets the coach's utterance - a \textit{parody} of a reminder - as a real reminder, which tends to repeat again; thus, the system na\"ively copies it. In the second example, the system needs to see more context to understand that the trip is an explanation from the patient for not making progress,  which the coach asked about in the previous utterance.  A better response should address the patient's explanation while maintaining specificity on the trip scenario, e.g., \textit{"No problem. Welcome back."}   


\section{Conclusions and Future Work}

We built an efficient health coaching dialogue system that helps patients create and accomplish specific goals, with a simplified NLU and NLG framework combined with mechanism-conditioned empathetic response generation. The experiments show that the system can generate more coherent and empathetic responses, supporting health coaches and improving coaching efficiency. In addition, our system outperforms the state-of-the-art in NLU tasks while requiring fewer annotations. We view our approach as a key step towards building automated and more accessible health coaching systems in low-resource settings and believe our approach may also generalize to building dialogue systems in similar scenarios, such as patient education at discharge or consulting on behavior change problems. 

In the future, we will explore the following directions: (1) Modelling empathetic understanding and generation with the goal response generator as one integrated end-to-end system while providing explainability. (2) A more comprehensive human evaluation from both the coach's and the patient's perspectives, including but not subject to goal completion and activity engagement rate.    
\section*{Acknowledgments}
This work is supported by the National Science Foundation under Grant IIS-1838770.


\bibliographystyle{acl_natbib}
\DeclareRobustCommand{\disambiguate}[3]{#3}
\bibliography{anthology,custom}

\appendix

\section{Training Details (COLING'22)}
All the following models use Huggingface Transformers Library ~\citep{hugginghttps://doi.org/10.48550/arxiv.1910.03771}. The hyperparameters are not extensively fine-tuned.
\begin{itemize}
\item \textbf{Slot-filling.} We use BERT-base as model backbone and associated tokenizer, with max sequence length of the tokenized input set to 50. The model was trained for \{\textbf{5.0}, 7.0, 10.0\} epochs by Adam with a learning rate of \{2e-5, \textbf{5e-5}\}, batch size of \{\textbf{32}, 64\}.
\item \textbf{Carryover Classifier.}
 We use BERT-base as model backbone and associated tokenizer, with max sequence length of the tokenized input set to 96. The model was trained for \{5.0, \textbf{7.0}, 10.0\} epochs by Adam with a learning rate of \{2e-5, \textbf{5e-5}\}, batch size of \{\textbf{16}, 32, 64\}.
\item \textbf{\texorpdfstring{NLG\textsubscript{hc}}{NLG hc}}
We use T5-base as model backbone and associated tokenizer, with max sequence length set to 128. The model was trained for 10.0 epochs by AdamW with a learning rate of 1e-4, warm up steps = 400, batch size of 64. We use sampling during decoding with top-k set to 50, top-p set to 0.95.

\item \textbf{Empathetic Generation}
We use GPT2 as model backbone and associated tokenizer, with max sequence length set to 96. The model was first trained for 10.0 epochs on the ED dataset by Adam with a learning rate of 1e-4, warm up steps = 400, batch size of 32. We use sampling during decoding with top-k set to 50, top-p set to 0.95. Then, the model was fine-tuned on 64 examples of health coaching empathetic data by 1 epoch. When decoding at inference, we use sampling with top-k set to 50, top-p set to 0.95.

\item \textbf{Emotion Detection}
 We use BERT-base as model backbone and associated tokenizer, with max sequence length set to 96. The model was trained for 8.0 epochs by Adam with a learning rate of 4e-5, and batch size of 32.
 
\end{itemize}
\section{Health Coaching Dataset Slot Examples}



Table~\ref{tab:the10} shows the description of the ten slots with value examples.

\begin{table*}[ht]
\centering
\begin{adjustbox}{width=\textwidth}

\begin{tabular}{lll}
Slot        & Description                           & Value Examples                           \\
\hline
\texttt{[activity]}    & The type of activity the patient will perform.                  & \textit{"walk, jogging, stair climbing"}            \\
\texttt{[amount]}       & The quantity of activity.              & \textit{"2000 steps, 6 fligths"}                    \\
\texttt{[duration]}    & The duration of the activity.          & \textit{"20 min, half an hour"}                     \\
\texttt{[distance]}     & The distance of the activity.          & \textit{"3 blocks, 2 miles, from home to bus stop"} \\
\texttt{[time]}        & The time of the day for the activity. & 
\textit{"at noon, after lunch, 4 pm"}               \\
\texttt{[location]}   & The location of the activity.          & \textit{"at work, at home, around the park"}        \\
\texttt{[dayname]}    & The days to do the activity.           & \textit{"Monday, Tuesday"}                          \\
\texttt{[daynumber]}   & The number of days for the activity.   & \textit{"3 days, 5 days"}                                         \\
\texttt{[repeatation]}  & The frequency of activity.             & \textit{"twice a day, daily"}                                    \\
\texttt{[score]}        & The confidence or attainability score.                  & Range from {[}1,10{]}  \\
\hline
\end{tabular}
\end{adjustbox}
\caption{The slot-value schema used in~\citet{gupta1-etal-2020-human}.}
\label{tab:the10}
\end{table*}

\begin{table*}[ht]
\centering
\begin{adjustbox}{width=\textwidth}


\begin{tabular}{ll}
\hline
Example Utterances  & Top-2 Predicted Emotion Labels\\ \hline
\textit{Sorry I left my fitbit in the emergency room yesterday.}             & Guilty (0.506),  Ashamed (0.323)                            \\
\textit{I'm not feeling very well yesterday so I did not go out for a walk.} & Disappointed (0.305), Ashamed (0.174)                       \\
\textit{I reached 10k steps last week can you believe that?}                 & Surprised (0.548), Proud (0.229)                            \\
\textit{Ok.}                                                                 & Angry (0.127), Furious (0.059)                              \\
\textit{I want to walk 3000 steps today.}                                    & Hopeful (0.484), Confident (0.132)  \\
\hline
\end{tabular}

\end{adjustbox}
\caption{The predicted emotion labels with corresponding probabilities by emotion cue detection on the patient's utterances.}
\label{tab:ecue}
\end{table*}

\section{Evaluation Metrics Description}

\paragraph{Partial/Complete Match} If all the values are correctly recorded for a given slot, it is considered a complete match. If the values are partially correct, it is considered a partial match.

\paragraph{Goal Correctness@$k$} Computes the percentage of correctness over all the predicted goals of each week. A goal contains ten attributes (slot-values); if at least $k$ attributes are predicted correctly, the goal is regarded as correct. Goal Correctness@$k$ is trivially equal to 100\% when $k = 0$.

\paragraph{BLEU} BLEU score~\citep{bleu-papineni-etal-2002-bleu} measures the word-level overlap between the generated output and the gold reference response. 

\paragraph{BertScore} BertScore~\citep{bert-score} measures the semantic similarity between the generated output and the reference leveraging BERT contextual embeddings.

\paragraph{Fluency} We measure fluency as perplexity (PPL) of the generated response using a pre-trained GPT2 model that has not been fine-tuned for this task, following previous work~\citep{ppl-ma-etal-2020-powertransformer,emp-rewriting10.1145/3442381.3450097}.

\paragraph{Empathy} We train a standard text regression model based on BERT using the response posts and corresponding level of empathy scores in  \citet{emp-type-data-sharma-etal-2020-computational} (achieving an RMSE of $\sim$0.57). We use this model to evaluate the empathy in the generated output compared with baselines.  

\section{Supplementary Analysis}

\subsection{Carryover Ablation}

Table~\ref{tab:carryab} shows the model performance of the carryover classifier with ablations. Using a combination of phases and acts can slightly improve recall with a tradeoff of reducing precision. However, using the dialogue context alone suffices for carryover classification with less annotation cost.

\subsection{Emotion Cue Detection}

Table~\ref{tab:ecue} shows the predicted emotion labels with corresponding probabilities by emotion cue detection on some patient’s utterance examples. The model can reasonably detect emotion signals of the patient's utterance only trained on the ED dataset. 

\begin{table}[!ht]
\centering
\begin{adjustbox}{width=0.9\columnwidth}
\begin{tabular}{lllll}
\hline
Input              & P             & R             & F1            & Acc           \\ \hline
Context Only       & \textbf{0.91} & 0.86          & \textbf{0.89} & 0.88          \\
Context+Phase     & 0.88          & 0.87          & 0.88          & 0.87          \\
Context+Act        & 0.88          & 0.86          & 0.87          & 0.87          \\
Context+Phase+Act & 0.90          & \textbf{0.87} & \textbf{0.89} & \textbf{0.89} \\
\hline
\end{tabular}
\end{adjustbox}
\caption{Model performance of carryover classification with ablations.}
\label{tab:carryab}
\end{table}

\end{document}